\documentclass{article}

\usepackage[preprint]{neurips_2023}
\usepackage{graphicx}
\usepackage{amsmath} 
\usepackage{multirow}
\usepackage{xcolor}
\usepackage{hyperref}
\hypersetup{
    colorlinks=true,
    urlcolor=blue!60!black
}

\usepackage[utf8]{inputenc} 
\usepackage[T1]{fontenc}    
\usepackage{hyperref}       
\usepackage{url}            
\usepackage{booktabs}       
\usepackage{amsfonts}       
\usepackage{nicefrac}       
\usepackage{microtype}      
\usepackage{xcolor}         
\usepackage{enumitem}
\usepackage{parskip}  
\title{Mini-Omni: Language Models Can Hear, Talk While Thinking in Streaming}

\author{%
  Zhifei Xie$^\spadesuit$$^\clubsuit$\thanks{Equal contribution. Work done during Zhifei Xie's internship at Inspirai.} \\
  \texttt{xzf24@mails.tsinghua.edu.cn} \\
  \and
\textbf{Changqiao Wu$^\spadesuit$\footnotemark[\value{footnote}]}\\
\texttt{wuchangqiao@inspirai.com} \\
\and
  $^\spadesuit$ Inspirai \quad
 $^\clubsuit$ Tsinghua University \\ 
 \href{https://github.com/gpt-omni/mini-omni}{\textcolor{blue!60!black}{\texttt{https://github.com/gpt-omni/mini-omni}}}
  }
\begin{document}
\maketitle




\begin{abstract}
Recent advances in language models have achieved significant progress. GPT-4o, as a new milestone, has enabled real-time conversations with humans, demonstrating near-human natural fluency. Such human-computer interaction necessitates models with the capability to perform reasoning directly with the audio modality and generate output in streaming. However, this remains beyond the reach of current academic models, as they typically depend on extra TTS systems for speech synthesis, resulting in undesirable latency. This paper introduces the \textbf{Mini-Omni}, an audio-based end-to-end conversational model, capable of real-time speech interaction. To achieve this capability, we propose a text-instructed speech generation method, along with batch-parallel strategies during inference to further boost the performance. Our method also helps to retain the original model's language capabilities with minimal degradation, enabling other works to establish real-time interaction capabilities. We call this training method \textbf{"Any Model Can Talk"}. 
We also introduce the \textbf{VoiceAssistant-400K} dataset to fine-tune models optimized for speech output. To our best knowledge,  \textbf{Mini-Omni} is the first fully end-to-end, open-source model for real-time speech interaction, offering valuable potential for future research.
\end{abstract}

\begin{figure}[h]
\begin{center}
    \includegraphics[width=0.8\textwidth]{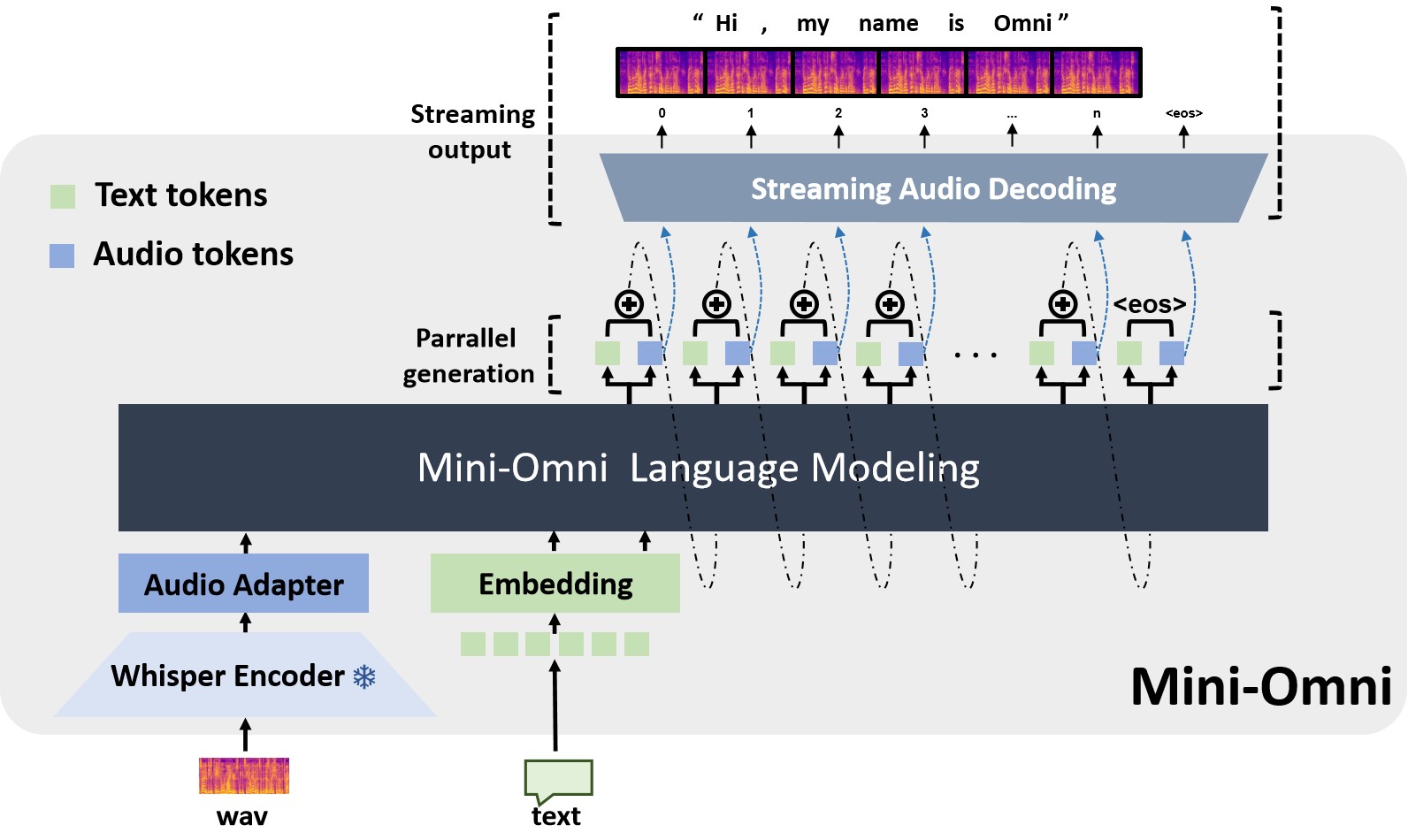}
\end{center}
\caption{The \textbf{Mini-Omni} model architecture.}
\end{figure}

\section{Introduction}
Recent developments in large language models have progressed rapidly, with models becoming increasingly powerful, such as off-the-shelf Llama 3.1 \citep{llama3.1}, Mixtral \citep{mixtral}, Qwen-2 \citep{qwen2}, and the well-known GPT-4. As an extension of their capabilities, language models are beginning to master understanding other modalities, exemplified by LLaVA \citep{llava}, Qwen2-Audio \citep{qwen2audio} and Video-llama \citep{videollama}. Despite their strength in specific tasks, a significant gap remains that hinders further integration of large language models into daily application: real-time voice interaction capability. GPT-4o \citep{gpt4}, introduced by OpenAI, is the first model to feature real-time multimodal speech interaction capabilities. It can understand and engage with vision, audio, and text while enabling real-time speech conversations, although it remains closed-source. Other models typically adopt two approaches to incorporate speech capabilities. The first is a cascade method, where the language model generates text, followed by a text-to-speech (TTS) model for audio synthesis. This approach introduces significant latency due to the time required for text generation, severely impacting user experience. The second, an end-to-end method like SpeechGPT \citep{speechgpt}, generates text before continuing to generate audio. However, this still requires waiting for text generation. Large language models need real end-to-end speech output capabilities to provide real-time feedback.

Enhancing models with speech output capabilities is a challenging task, primarily due to four factors: (1) \textbf{Complexity of Audio Reasoning}: Our experiments indicate that direct training for audio modality reasoning is highly challenging, often resulting in incoherent outputs from the model. (2) \textbf{Model Complexity}: Incorporating additional modules for speech input and output increases the overall complexity. (3) \textbf{Difficulty in Modality Alignment}: The reasoning abilities developed for text are difficult to transfer to the audio domain. (4) \textbf{Resource Demands}: Adapting a model's text capabilities to the speech modality requires converting all data labels into audio and retraining, significantly increasing resource consumption.

In this paper, we propose \textbf{Mini-Omni}, the first open-source multi-model large language model with real-time conversational capabilities, featuring fully end-to-end speech input and output abilities. It also includes various other audio-to-text functionalities such as Automatic Speech Recognition (ASR). We adapt currently available off-the-shelf methods for discretizing speech tokens and employ the simplest model architecture, making it easy for our model and approach to be adapted by other researchers. 
Direct audio reasoning poses significant challenges; however, our approach successfully addresses this using only a 0.5B model and a limited amount of synthesized audio data. Importantly, our training framework achieves this without heavy reliance on extensive model capabilities or large volumes of data.

To leverage and preserve the original capabilities of the language model, we propose a parallel generation paradigm in which the transformer simultaneously produces audio and text tokens. Subsequently, we observed a minimal impact of the audio modality on text capabilities and further introduced \textbf{batch-based parallel generation}, which significantly enhances the model’s reasoning ability during streaming audio output. As a poinerr, we opted not to sacrifice audio quality for a simpler and lower bitrate audio encoder, in order to reduce the complexity of audio inference in the model. However, to ensure audio quality, we selected SNAC \citep{snac}, a music-grade encoder features 8 layers of codebooks and processes hundreds of tokens per second. Innovatively, we applied \textbf{text-instructed delayed parallel generation} to address the issue of long SNAC codebook sequences. Experiments show that the audio output quality is on par with common TTS systems.

We also propose a method that requires minimal training and modification of the original model, enabling other works to rapidly develop their own speech capabilities. We refer to this approach as \textbf{"Any Model Can Talk"}, designed to achieve speech output using a limited amount of additional data. The approach extend speech capabilities through additional adapters and pre-trained models, fine-tuning with a small amount of synthesized data. This is combined with the aforementioned parallel modeling approach to enable streaming output in the new modality while preserving the original model’s reasoning capabilities. 

To evaluate the capabilities of \textbf{Mini-Omni}, we first assessed its performance on traditional text-to-speech multi-modal tasks, including text-based question answering (textQA), automatic speech recognition (ASR), text-to-speech response, and speech-based question answering (speechQA). The model demonstrated strong proficiency in these fundamental tasks. Additionally, we conduct a series of experiments to investigate the impact on the original model's capabilities and assess the effectiveness and variations of our inference method. Preliminary experiments demonstrate that batch parallel inference preserves the model’s original capabilities. We will conduct further experiments and provide additional details in due course.

Lastly, we observed that most open-source QA datasets contain mixed code or overly lengthy text, rendering them unsuitable for speech model. To overcome this limitation, we introduce the \textbf{VoiceAssistant-400K} dataset, comprising over 400,000 entries specifically generated by GPT-4o for speech assistant supervised fine-tuning (SFT).

\noindent \textbf{In summary, we make the following contributions:}
\begin{itemize}[leftmargin=*]
    \item We introduce \textbf{Mini-Omni}, the first open-source end-to-end multimodal large model with audio input and audio streaming output capabilities. We propose a unique text-instruct parallel generation method that enables speech inference outputs aligned with textual capabilities, achieved with minimal data. We further enhance this with delayed parallelism, accelerating audio inference speed. 
   
    \item We introduce "\textbf{Any Model Can Talk}", an innovative approach that enhances performance without altering the architecture of large models by focusing on training and inference. Our method employs a three-phase training process for speech-to-text and text-to-speech adapters, including annealing and SFT. Our method involves minimal training and modification of the original model, aiming to provide a reference for incorporating interaction capabilities into other models.
    
    \item We identified shortcomings in existing open-source QA datasets when training audio assistants and proposed a dedicated dataset for speech model outputs, called\textbf{ VoiceAssistant-400K}. This dataset, synthesized using GPT-4o, can be used to fine-tune models to develop the tone of a voice assistant.
\end{itemize}

\section{Related Work}
\textbf{Multimodal Understanding} Recently, researchers have been increasingly focused on advancing unified models for cross-modal understanding. These approaches typically employ a well-pretrained neural network as the encoder for relevant modalities, using a lightweight adapter to align the encoder's output with the text input of language model. Classical works such as LLaVA \citep{llava}, Flamingo \citep{flamingo} and BLIP \citep{blip} are used for visual understanding, while in the audio domain, models like Whisper \citep{whisper} and Beats \citep{beats} are commonly utilized as encoders for semantic and acoustic features. In Llama 3.1, Whisper is employed, while SpeechVerse \citep{speechverse} leverages WavLM \citep{wavllm}; SALMONN \citep{salmonn}, combine Whisper and Beats to extract features. Such works are often constrained to producing output in the text modality.

\textbf{Audio Language Modeling} Recently, an increasing number of studies have employed audio tokenization to bridge the gap between audio and text. Audio tokenization converts continuous audio signals into discrete audio tokens, enabling large language models to perform inference and even cross-modal interactions. As a result, a variety of speech-text tasks, such as ASR, TTS, music understanding and generation, and sound editing, can be accomplished. MegaTTS \citep{megatts} utilized audio codecs for speech synthesis, while efforts like InstructTTS \citep{instructtts}, SpearTTS \citep{spearTTS}, and Voicebox \citep{voicebox} have further explored optimizations in decoding methods and conditioning techniques, employing Diffusion as the converter from tokens to audio.

\textbf{Real-Time Human-Machine Interaction Models} Since the introduction of GPT-4o \citep{gpt4}, real-time conversational models have achieved unprecedented results, providing near-instantaneous voice feedback to user inputs, marking a significant milestone for the next generation of multi-modal large models. However, the technical implementations remain proprietary. Models with real-time interaction capabilities are currently scarce. SpeechGPT \citep{speechgpt} is an early end-to-end speech interaction model; however, it still suffers from latency due to the Audio-Text-Text-Audio(A-T-T-A) process, similar to Spectron \citep{Spectron}. LauraGPT \citep{lauragpt} also employs a similar approach but not for voice conversation scenario. VITA \citep{vita} and Qwen-audio2 \citep{qwen2audio} are two models that support voice input, but they output text and rely on external TTS systems for speech synthesis. \textbf{Mini-Omni} is a fully end-to-end speech-to-speech conversational model. Through our exploration, we have identified the biggest challenge in advancing this field: the logical inconsistency in reasoning when only the audio modality is present, which we will address in the following chapter.

\section{Mini-Omni}

Our innovation stems from existing methods such as SpeechGPT \citep{speechgpt} and Spectron \citep{Spectron} utilize the A-T-T-A approach, which mitigates the challenges of direct audio learning by guiding the speech generation process through text. However, generating text first and then audio is suboptimal for real-time dialogue scenarios. To address this, we propose a novel method for simultaneous text and audio generation. This approach hypothesizes that text outputs have higher information density, allowing for the same response with fewer tokens. During the generation of audio tokens, the model effectively conditions on corresponding text tokens, akin to an online TTS system. Prior to generating audio tokens, padding with \( N \) tokens ensures that the corresponding text tokens are produced first, allowing this to serve as a hyperparameter adjustment. Additionally, the model can also condition on speaker and style embeddings, facilitating control over speaker characteristics and stylistic elements. In this section, we will detail how we implement our idea step by step.

\subsection{Audio Language Modeling}
Consider \( Y = (y_i \in \mathcal{V}_{\text{txt}} \mid i = 1, \ldots, t_{\text{txt}}) \) as a text utterance from a vocabulary \(\mathcal{V}_{\text{txt}}\) with length \( t_{\text{txt}} \). The probability of \( Y \) can be expressed as \( p(Y) = \prod_{i=1}^{t_{\text{txt}}} p(y_i \mid y_1, \ldots, y_{i-1}) \). Now, when dealing with a continuous speech signal, we can convert it into discrete speech tokens (\(\text{dst}\)), represented as \( D = (d_i \in \mathcal{V}_{\text{dst}} | i = 1, \cdots , t_{\text{dst}}) \) using a tokenizer. In this context \(\mathcal{V}_{\text{dst}}\) is the vocabulary of discrete speech tokens. These discrete speech tokens can be treated as spoken language within \(\mathcal{V}_{\text{dst}}\) and modeled in a manner similar to text. We combine text and speech in a new vocabulary \(\mathcal{V}_{\text{voxt}}\) by \( \mathcal{V}_{\text{voxt}} = \mathcal{V}_{\text{txt}} \cup \mathcal{V}_{\text{dst}} \). Therefore, we can model the probability of both speech and text tokens as \( Z \), where \( Z = (z_i \in \mathcal{V} | i = 1, \cdots , t) \). This probability is expressed as \( p(Z) = \prod_{i=1}^{t} p(z_i \mid z_1, \cdots, z_{i-1}) \), \( Z \) represent discrete speech tokens \( D(\mathcal{V} = \mathcal{V}_{\text{dst}}) \) or text tokens \( Y(\mathcal{V} = \mathcal{V}_{\text{txt}}) \) or various combinations of \( Y \) and \( D \). For the audio and text tokens generated simultaneously, the negative log-likelihood loss can be formulated as in Equation (1).
\begin{equation}
\mathcal{L}(T,A|C) = \sum_{j=1}^{m}\sum_{i=1}^{n_{j}} \log P(T_{i,j}, A_{i,j} | T_{<i,j}, A_{<i,j}; X_j)
\end{equation}

where $T$, $A$ is the text-audio output pairs in the training corpus $C$, and $m$ is the number of training examples. $X_j$ is the input condition of j-th example, $n_j$ is max number of tokens of sample $T_j$ and $A_j$, $T_{i,j}$ and $A_{i,j}$ represent the i-th text token and audio token of j-th sample.

\subsection{Decoding Strategies}

\textbf{Audio Generation with text instruction.} Language models have undergone substantial advancements, demonstrating exceptional reasoning capabilities within the text modality. In response, \textbf{Mini-Omni} has been restructured to transfer these reasoning abilities to streaming audio output through a text-audio parallel decoding approach. This method simultaneously outputs both audio and text tokens, with the audio generated via text-to-speech synthesis, ensuring real-time delivery while leveraging the text-based reasoning strengths. To align with the inputs of large models, all sequences generated in parallel are summed before producing the next token, as illustrated in Figure 1. This approach enables the model to achieve real-time voice output in chat scenarios with minimal first token delay.

\textbf{Text-delay Parallel Decoding.} Parallel generation was first introduced by MusicGen \citep{MusicGEN} to accelerate the music generation process, and we have integrated this approach into the text modality to enhance reasoning capabilities. Parallel decoding is feasible because audio token codebooks used in language model training typically consist of multiple layers; generating all layers simultaneously can significantly increase model speed. For real-time speech output models, parallel decoding is even more critical, allowing for the generation of hundreds of audio tokens per second on standard devices. In this paper, we employ SNAC as the audio encoder, which comprises seven token layers with complementary relationships. Therefore, we employ eight sub-Language Model heads to generate eight tokens, including text, in a single step, while maintaining a one-step delay between adjacent layers. Since audio tokens are derived from text synthesis, the text token is output first, followed by SNAC tokens from the first to the seventh layer. The process of text-first delay parallel decoding we propose is illustrated in Figure 2(b).
\begin{figure}[h]
\begin{center}
    \includegraphics[width=\textwidth]{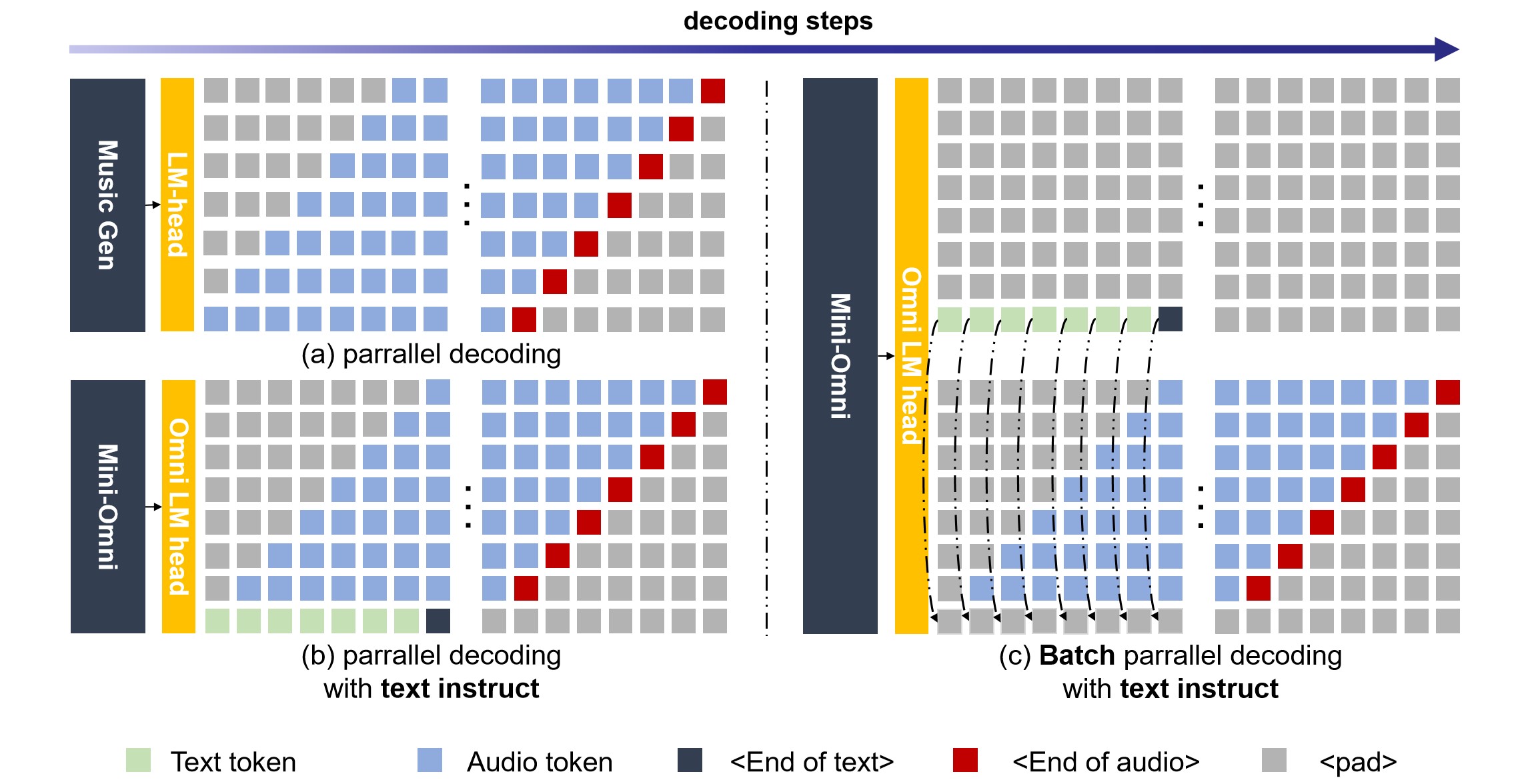}
\end{center}
\caption{\textbf{Mini-Omni} incorporates text-instruct mechanisms alongside Batch parallel generation techniques.
}
\end{figure}

\textbf{Batch Parallel Decoding.} Although the previously introduced parallel generation method effectively transfers reasoning capabilities from the text modality to the audio modality, our experiments reveal that the model's reasoning performance still varies between text and audio tasks, with audio responses tending to be simpler. We hypothesize that this is due to limitations in model capacity or insufficient audio data. To address this issue and further enhance the model's reasoning capabilities during dialogue, maximizing the transfer of its text-based abilities, we experimentally employ a Batch approach. Given the model's stronger performance in the text modality, we expand the inference task for a single input to a batch size of 2: one sample requires both text and audio responses, as described earlier, while the other sample only requires a text response, focusing on text-based audio synthesis. However, the text token output from the first sample is discarded, and the text output from the second sample is embedded into the corresponding text token positions of the first sample. Simultaneously, the audio from the first sample is streamed using the content from the text-only response of the second sample; we term this process batch parallel decoding. Through this method, we effectively and almost entirely transfer the model's text-based capabilities to the audio modality with minimal resource overhead, significantly enhancing its reasoning abilities in the new modality. The inference process of batch parallel decoding is illustrated in Figure 2(c). We believe batch parallel decoding represents a key algorithmic innovation that enables such a small model to exhibit strong conversational abilities.

\subsection{Any Model Can Talk}
In this section, we present our training methodology. Our approach is designed to preserve the capabilities of the original model as much as possible. This is achieved firstly due to the strong performance of our base model, and secondly because our method can be applied to other works that excel in text output but lack robust speech interaction capabilities.
\begin{figure}[h]
\begin{center}
    \includegraphics[width=0.9\textwidth]{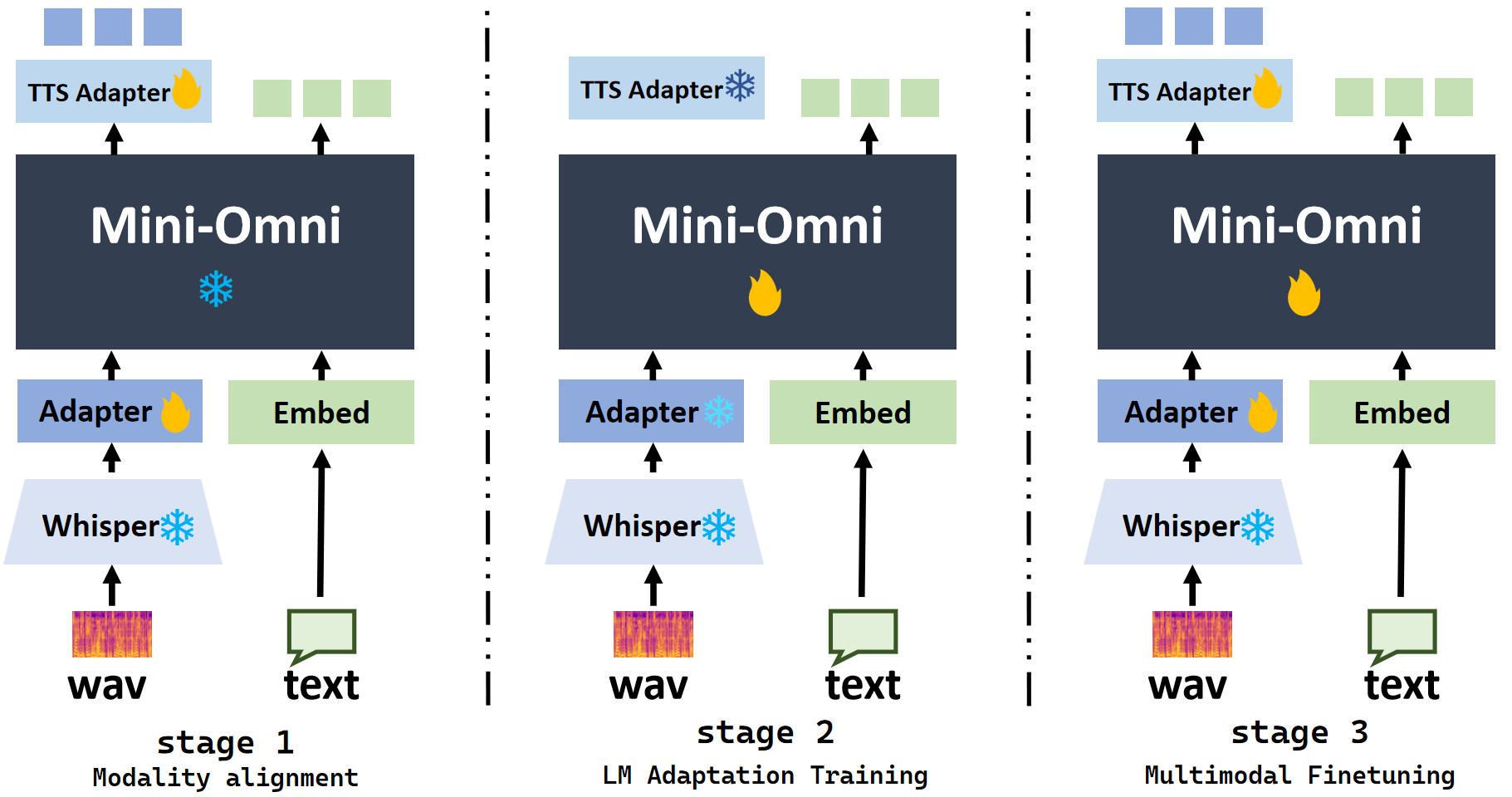}
\end{center}
\caption{Mini-Omni's three-stage training phases: modality expansion, modality adaptation training, and holistic fine-tuning.}
\end{figure}

\textbf{Audio Encoding}: The audio input primarily focuses on feature extraction from the input audio, with options including Hubert or a separately pretrained audio encoder. Given our focus on speech input, Whisper \citep{whisper} and Qwen2-audio \citep{qwen2audio} also demonstrate effective performance for general audio tasks. For audio output, selecting audio tokens with a multi-codebook approach better captures audio details. We experimented with flattening for audio token modeling, but it resulted in excessively long tokens, which are detrimental to streaming and lead to unstable learning. Instead, parallel decoding, inspired by MusicGen \citep{MusicGEN}, employs a delay pattern combined with text conditions, as illustrated in Figure 2.

\textbf{Three-Stage Training. }Our training methodology is divided into three distinct stages: \textbf{(1) Modality Alignment.} The goal of this stage is to enhance the text model's ability to understand and generate speech. The core model of \textbf{Mini-Omni} is entirely frozen, with gradients allowed only in two adapters. During this stage, we use data from speech recognition and speech synthesis to train the model's speech recognition and synthesis capabilities. \textbf{(2) Adaption Training.} Once the new modality is aligned with the text model's input, the adapters are frozen. In this stage, we focus solely on training the model's text capabilities when given audio inputs, as audio output is simply synthesized from text. The model is trained using data from speech recognition, spoken question answering, and text response tasks. \textbf{(3) Multi-modal Finetuning.} In the final stage, the entire model is fine-tuned using comprehensive data. At this point, all model weights are unfrozen and trained. Since the primary modality alignment tasks are handled during adapter training, the original model's capabilities are maximally preserved.

\textbf{Model Input Ids.} Given the eight parallel output sequences, the input also requires eight sequences, leading to significant complexity. Therefore, we briefly outline the organization of model inputs here. The model can accept either text or audio inputs, which are placed in the corresponding modality sequences. For audio inputs, the input tokens and Whisper features are transformed into tensors of the same dimension via adapters and then concatenated. Depending on the task, we place the <answer> special token in different positions to guide the model's output, achieving multi-modal output. The organization of some tasks is illustrated in Figure 4. Before being fed into the model, all sequences are summed and averaged to integrate features.
\begin{figure}[h]
\begin{center}
    \includegraphics[width=\textwidth]{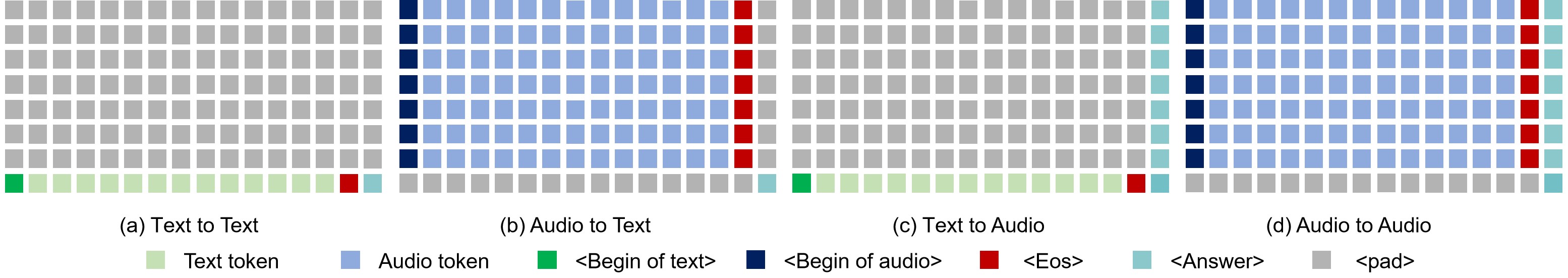}
\end{center}
\caption{Diagram of the input section of \textbf{Mini-Omni} parallel generation. The \textbf{<answer>} special token is placed at the end of the sequence to be generated, as determined by the task.}
\end{figure}

\section{Experiments}
his section presents the foundational capability test results for \textbf{Mini-Omni}. We first describe the training datasets, data processing methods, and hyperparameters. We then evaluate the model's performance on core tasks like speech recognition and provide several use case examples. 
We will include all relevant experiments in the next version as soon as possible.

\subsection{Datasets}

To establish foundational speech capabilities, we trained the model using three speech recognition datasets totaling approximately 8,000 hours, focusing on speech understanding and synthesis. For text modality, we incorporated 2 million data points from the Open-Orca \citep{openorca} dataset and integrated them with other modalities to preserve textual accuracy. Moss's SFT dataset \citep{moss} was utilized with zero-shot TTS to synthesize 1.5 million speech QA pairs. To avoid unsuitable code and symbolic outputs, we created the VoiceAssistant-400K dataset with GPT-4o. Datasets are detailed in Table 1. Stage 1 involves ASR data for training speech adapters. Stage 2 uses TextQA and AudioQA for audio/text input and text response training. Stage 3 focuses on multimodal interaction using the audio modality of AudioQA. Final stage training includes annealing and fine-tuning with Voice QA.
\begin{table}[h]
\centering
\begin{tabular}{lllll}
\toprule
\textbf{Task} & \textbf{Stages}   & \textbf{Dataset} & \textbf{Modality} & \textbf{items} \\ 
\midrule
                                 &   & Libritts \citep{libritts}  & A1|T1 & 586 h \\
  ASR & 1,2,3 & VCTK \citep{vctk} & A1|T1 & 44 h  \\
                                  & & Multilingual LibriSpeech \citep{mls}  & A1|T1 & 8000h \\ \midrule
 Text QA & 2,3& Open-Orca \citep{openorca}  & T1|T2 & 2000K  \\ \midrule
 Audio QA & 3 & Moss-002-sft-data \citep{moss} & A1|T1|A2|T2 & 1500K \\ \midrule
          & & Alpaca-GPT4 \citep{Alpacagpt4}  & A1|T1|A2|T2 & 55k \\
           & & Identity finetune \citep{identity_finetune} & A1|T1|A2|T2 & 2k \\
           & & QAassistant \citep{qa-assistant-2} & A1|T1|A2|T2 & 27k \\
  voice QA  & final & Rlhf \citep{rlhf} & A1|T1|A2|T2 & 367k \\ 
        & & Trivia-singlechoice \citep{t-sim}  & A1|T1|A2|T2 & 17k \\
       &  & Trivia-Multichoice \citep{t-mul}  &A1|T1|A2|T2 & 20k \\
        &  &  OpenAssistant \citep{open-assis} & A1|T1|A2|T2 & 2k \\
\bottomrule
\end{tabular}
\caption{The datasets and their usage for training \textbf{Mini-Omni} are as follows: In the modality notation, T and A represent the text and audio modalities, with subscripts 1 and 2 indicating input or output.}
\end{table}
\subsection{Training Parameters} 
Our model is trained on 8 A100 GPUs, utilizing a cosine annealing learning rate scheduler with a minimum learning rate of 4e-6 and a maximum learning rate of 4e-4. Each training epoch consists of 40,000 steps, with batch size 192 for each step. The base language model employs Qwen2-0.5B \citep{qwen2}, a transformer architecture with 24 blocks and an internal dimension of 896. The speech encoder uses the Whisper-small encoder, with ASR adapter connected via two-layer MLP, and the TTS adapter extends the original model by adding 6 additional transformer blocks. During fine-tuning, we use learn rate from 4e-6 to 5e-5.

\subsection{Experimental Results} 
We first evaluated the model's performance on ASR tasks to assess its speech understanding capabilities. Basic experiments on speech recognition capabilities were conducted using the four test sets from LibriSpeech \citep{librispeech}: test-clean, test-other, dev-clean, and dev-other. Results are presented in Table 2, where we compare the accuracy of our adopted speech recognition systems, wav2vec2 \citep{wav2vec} and Whisper-small, as well as the VITA \citep{vita}. The findings indicate that while Mini-Omni's speech recognition performance slightly lags behind Whisper-small’s \citep{whisper} decoder, it still achieves an excellent level of audio comprehension.
\begin{table}[h]
\centering
\begin{tabular}{lcccc}
\toprule
\textbf{Method} & \textbf{test-clean} & \textbf{test-other} & \textbf{dev-clean}  & \textbf{dev-other} \\ 
\midrule
wav2vec2-base \citep{wav2vec}  &  6.0 &13.4 & - & - \\
VITA \citep{vita} &  8.14 & 18.41 & 7.57 & 16.57  \\
whisper-small  \citep{whisper} & 3.4  & 7.6 &  - & - \\
\textbf{Mini-Omni} &  4.5 & 9.7 & 4.6 &  9.2  \\
\bottomrule
\end{tabular}
\caption{Comparison of the model's ASR with the base model used.}
\end{table}

\subsection{Case Study}
Here, we present several cases to demonstrate \textbf{Mini-Omni}'s capabilities in speech understanding and reasoning. These examples reveal that speech-based reasoning is somewhat weaker compared to text-based reasoning, highlighting the necessity for batch generation. For more impressive examples, please refer to https://github.com/gpt-omni/mini-omni.
\begin{figure}[h]
\begin{center}
    \includegraphics[width=0.9\textwidth]{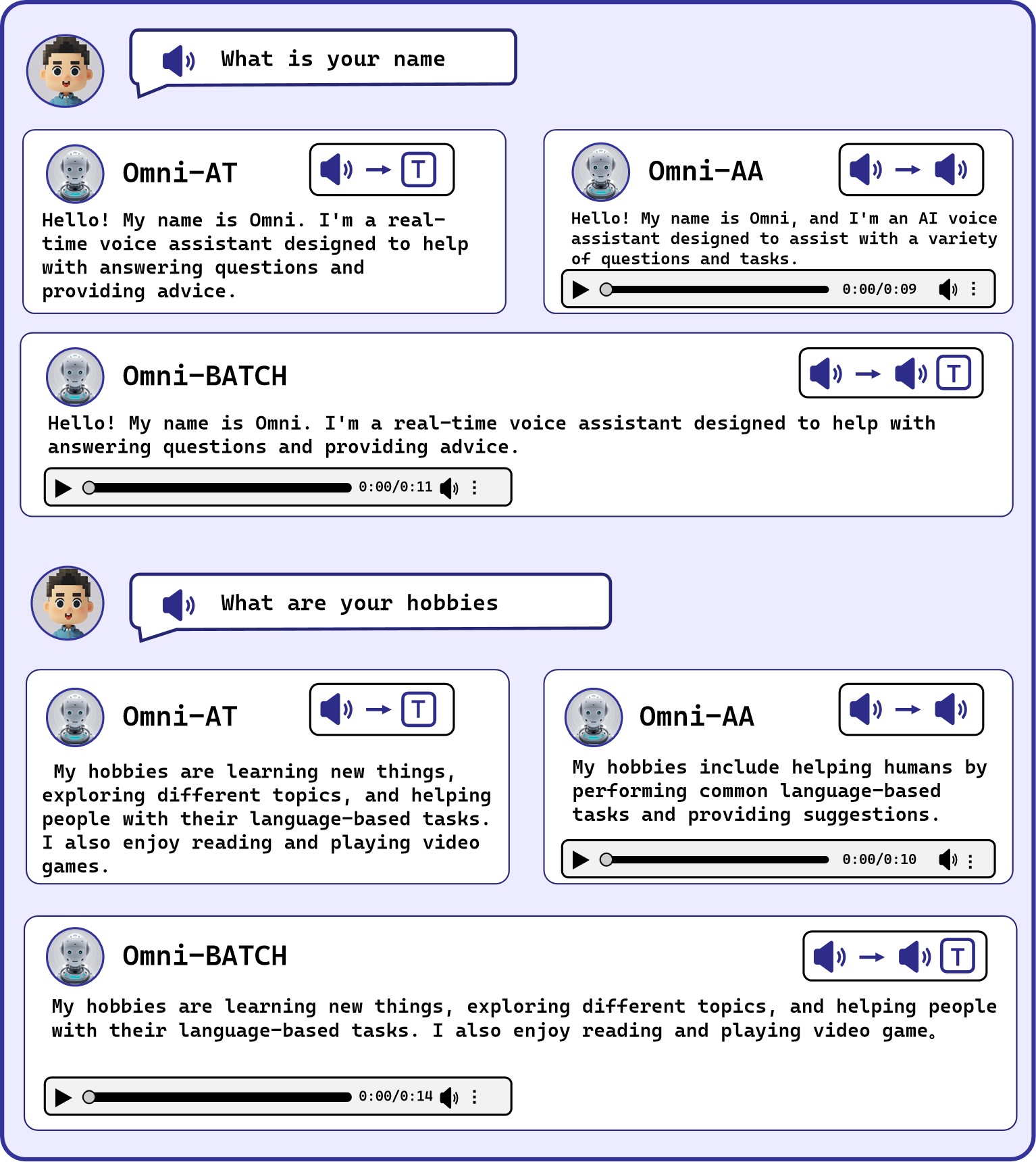}
\end{center}
\caption{Real streaming output examples of Mini-Omni}
\end{figure}

\section{Conclusion}
In this work, we introduce \textbf{Mini-Omni}, the first multi-modal model with direct speech-to-speech capabilities. Building on previous approaches that use text-guided speech generation, we propose a parallel text and audio generation method that leverages minimal additional data and modules to rapidly transfer a language model's text capabilities to the audio modality, supporting streaming output interactions with high model and data efficiency. We explore both text-instructed streaming parallel generation and batch parallel generation, which further enhance the model's reasoning ability and efficiency. Our approach successfully addresses challenging real-time dialogue tasks using a model with only 0.5 billion parameters. We have developed the \textbf{Any Model Can Talk} method, based on a pre and post-adapter design, to facilitate rapid speech adaptation of other models with minimal additional training. Additionally, we have released the VoiceAssistant-400K dataset for fine-tuning speech output, designed to minimize the generation of code symbols and assist humans in a voice assistant-like manner. All our data, inference, and training codes will be progressively open-sourced at \url{https://github.com/gpt-omni/mini-omni}. We hope to provide guidance and support for other work focused on language model speech interaction.

\bibliographystyle{plainnat} 
\bibliography{ref}
\end{document}